# Structure design and coordinated motion analysis of bionic crocodile robot

Jun Wang, Jingya Zheng, Yuhang Zhao, Kai Yang*

*Abstract*—Crocodiles, known as one of the oldest and most resilient species on Earth, have demonstrated remarkable locomotor abilities both on land and in water, evolving over millennia to adapt to diverse environments. In this paper, we draw inspiration from crocodiles and introduce a highly biomimetic crocodile robot equipped with multiple degrees of freedom and articulated trunk joints. This design is based on a comprehensive analysis of the structural and motion characteristics observed in real crocodiles. The bionic crocodile robot has the problem of limb-torso incoordination during movement, in order to solve this problem, we apply the D-H method for both forward and inverse kinematics analysis of the robot's legs and spine. Through a series of simulation experiments, we investigate the robot's stability of motion, fault tolerance, and adaptability to the environment in two motor pattern: with and without the involvement of the spine and tail in its movements. Experiment results demonstrate that the bionic crocodile robot exhibits superior motion performance when the spine and tail cooperate with the extremities. This research not only showcases the potential of biomimicry in robotics but also underscores the significance of understanding how nature's designs can inform and enhance our technological innovations.

## I. INTRODUCTION

The crocodile is a highly evolved reptile that can perform almost all the quadrupedal gaits of mammals and has a powerful tail. The structural features of the crocodile [1] make it well-adapted to its environment. Its great adaptability to the environment is achieved through the coordination between the legs and other parts of the body, such as the trunk, head, and tail. Crocodiles extend their stride length by bending their bodies; they achieve balance by wagging their tails. These behaviors suggest that limb coordination plays an important role in the control of biological locomotion [2-4], as they adapt to the environment through the coordination of multiple parts and allow a single part to perform multiple motion functions with the cooperation of other parts. However, there is still a lot of room for research on this limb-body coordination mechanism, and the study and improvement of this mechanism can help elucidate the motor control of crocodilians and also be useful in the design of bionic crocodilian robots.

At present, there are few studies using alligators as robotic bionic objects all over the world. A bionic crocodile modular robot consisting of 14 small robot modules was designed at Ohio State University, USA, using an American alligator as a bionic object [5, 6]. A low-cost open-source bionic crocodile robot platform was designed by a research team at the Indian Institute of Technology [7]. This platform allows rapid prototyping of robots and facilitates iterative design. Based on the bionic crocodile robot platform, the team investigated the effect of robot body torso swing on robot motion [8]; a research team at BITS Pilani (India) designed a new modular robot 2DxoPod by imitating the motion of vertebrates such as crocodiles and snakes [9]. This modular robot was designed with two mutually overlapping and orthogonal degrees of freedom to imitate the joints of creatures such as snakes, dogs, and crocodiles, and the overall design was optimized by reducing the number of drives, degrees of freedom, and coordinates of the robot during navigation. However, none of these studies addressed the structural design of the crocodile spine and tail and their role in locomotion.

This paper draws inspiration from the remarkable characteristics of crocodiles and applies them to the design of a bio-inspired crocodile robot. The research encompasses a comprehensive structural analysis, the kinematic model established using D-H method-, and both forward and inverse kinematic analyses. The research focuses on achieving coordinated planning control among the various components of the bio-inspired crocodile by extracting the relationships between limb phase, spine angles, and tail angles during a single movement cycle, closely mimicking the locomotion of real crocodiles. Subsequently, experimental analysis is conducted to evaluate several key aspects of the bio-inspired crocodile's performance. This assessment includes an examination of its motion stability when subjected to coordinated planning control of the spine, tail, and limbs, an exploration of its resilience in the face of potential component damage, and an investigation into its adaptability within different environment.

Overall, this research not only highlights the potential of bio-inspired robotics but also provides insights into the robustness and adaptability of the bio-inspired crocodile robot in various environment.

## II. STRUCTURAL DESIGN

### A. Crocodile and its movement characteristics: the biomimetic inspiration

Crocodiles are remarkable semi-aquatic and semi-terrestrial creatures that inhabit a wide range of regions across the world, spanning tropical, subtropical, and temperate areas in Asia, Africa, America, and Oceania. Globally, there

*Research supported by Jiangsu Graduate Research And Innovation Program (KYCX21_2251).

Jun Wang is with the China University of Mining and Technology, Xuzhou, CO 314001 China (e-mail: jrobot@126.com).

Jingya Zheng is with the China University of Mining and Technology, Xuzhou, CO 314001 China (e-mail: jyzheng99@163.com).

Yuhang Zhao is with the China University of Mining and Technology, Xuzhou, CO 314001 China (e-mail: zhao20232022@163.com).

Kai Yang is with the China University of Mining and Technology, Xuzhou, CO 314001 China (corresponding author: e-mail: yk267x@163.com).

are currently 26 recognized crocodile species, all belonging to the biological order Alligatoridae [10-13]. Within the Alligatoridae order, crocodiles are categorized into three major families: Alligatoridae, Crocodylidae, and Gavialidae. These diverse species exhibit a vast range of sizes, with the African Nile Crocodile, the largest among them, reaching an average length of approximately 4 meters [14]. In contrast, the Chinese alligator (Alligator sinensis), a member of the Alligatoridae family, represents the smallest crocodile species, typically measuring around 1.25 meters in length [15].

In this section, we delve into an analysis of crocodiles within the Alligatoridae family, focusing on an examination of their structural characteristics. Additionally, Fig.1 illustrates the skeletal structures of Alligatoridae crocodiles.

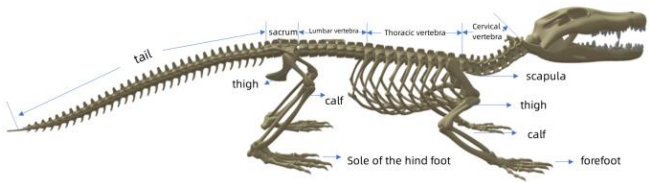

Figure.1 Skeletal structure of the alligator

An intriguing observation reveals that the crocodile's tail boasts a greater number of segments than the trunk, with the tail's length accounting for roughly half of the crocodile's overall body length. Crocodile legs are relatively short, with the scapula connecting the front two legs. The front legs exhibit five toes, whereas the hind legs feature four toes, and the hind feet adopt a webbed structure.

Through continuous evolution, crocodiles have acquired exceptional mobility and adaptability, demonstrating peak performance in various modes of movement, whether in water or on land. Common locomotion patterns for crocodiles include Belly Crawl, High Walk, Galloping, swimming, and underwater rolling, among others [16].

Belly Crawl, the most frequently observed terrestrial locomotion mode, involves the crocodile's body remaining in close proximity to the ground with minimal movement. During this mode, the front and hind legs move in a diagonal gait, while the tail swings alternately from side to side, creating a leisurely pace for crocodile movement.

On the other hand, sees crocodiles raising their legs upright beneath their bodies, with their feet aligned in the direction of movement. They move in a diagonal gait, and about half of their tail alternates in swinging along the ground. This mode typically ranges in speed from 5 to 10 km/h.

Additionally, crocodiles, with their impressive burst of power, are capable of Galloping movements, which represent their fastest mode of locomotion and are often employed when attempting to escape. In aquatic environments, crocodiles tightly press their limbs against their body's sides while using their tail to generate the primary thrust for swimming. The tail's swinging motion resembles a sine wave, and crocodiles use rapid tail swings to accelerate. During this movement, the body trunk exhibits wave-like undulations resembling a sine curve [17].

## B. Structural design

In this study, we have selected the spectacled caiman, a member of the Alligator genus Alligator, as our research subject. The design of the bionic crocodile robot was inspired by the distinct morphological characteristics of this species. The robot comprises multiple components, including the head, body trunk, limbs, and tail, with each part consisting of interconnected joints. For a visual representation of the overall structure, refer to Fig.2.

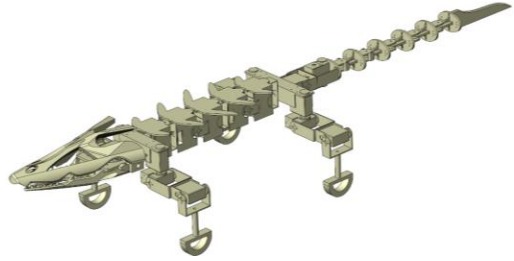

Figure.2 design of the overall structure of the bionic crocodile robot

The leg structure of the bionic crocodile robot was developed through an analysis of the leg skeleton structure and movement patterns [18] of real crocodiles, as depicted in Fig. 3, two degrees of freedom have been set in the leg, including hip-joint and knee-joint. When moving on a flat surface, the robot adopts the crawling structure illustrated in Fig. 3(a). On the other hand, when traversing rough terrain, it utilizes the lactation type structure presented in Fig. 3(b). This design allows the bionic crocodile robot to dynamically switch between different structures and gaits based on the encountered environment.

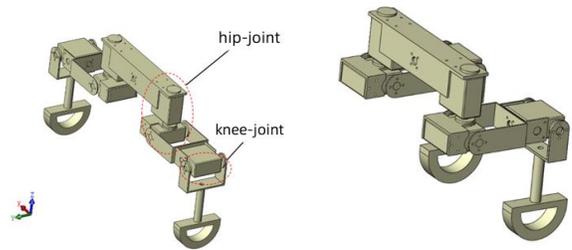

Figure.3 Design of leg structure of bionic crocodile robot

In this paper, the torso part of the bionic crocodile robot is designed according to the body structure and motion of the real crocodile torso [19]. As shown in Fig.4, five degrees of freedom have been set in the torso, including three degrees of freedom for the lateral movement of the spinal column and two degrees of freedom for the robot's pitch motion.

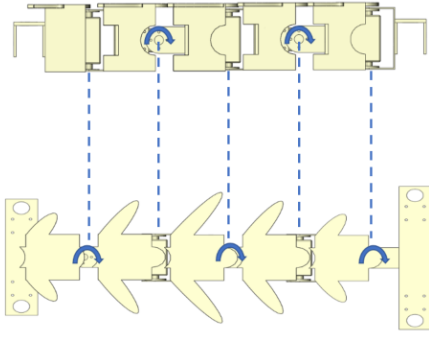

Figure.4 The design of the torso of the bionic crocodile robot

The design of the tail structure [20] of the crocodile robot is shown in Fig.5. As shown in Fig.5, the tail of the bionic alligator robot consists of a drive section and an under-drive section, where the drive section has six degrees of freedom and moves with a linear drive; the under-drive section is made of flexible material, which can more realistically simulate the flexible tail of the alligator.

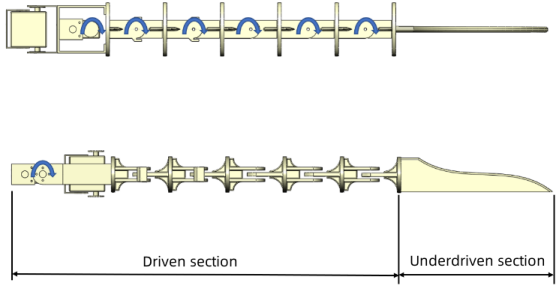

Figure.5 The design of the tail of the bionic crocodile robot

## III. KINEMATIC ANALYSIS

The appropriate coordinate system is set for each joint to describe the correct kinematic characteristics of the bionic crocodile robot. In this paper, the reference coordinate system [21] of the legs and spine of the bionic crocodile robot [22] was constructed by using D-H representation, the homogeneous transformation matrix between the coordinate system of the robot's foot movement and the reference coordinate system of the base was obtained, and the relevant kinematic model was established.

### A. Kinematic model of legs

Taking the example of the left hind leg of the bionic crocodilian robot, this paper establishes the coordinate system of each rod and derives the corresponding kinematic equations by the D-H method.

The D-H linkage coordinate system for the leg of the bionic crocodile robot according to the above steps is shown in Fig.6.

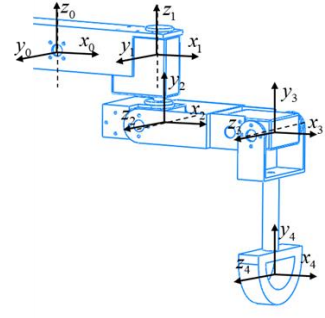

Figure.6 D-H coordinate system of the hind limb

According to the chain rule of coordinate system transformation, the flush transformation matrix of adjacent joint coordinate systems is expressed as:

$$^{n-1}T_n = \begin{bmatrix} C\theta_n & -S\theta_n & 0 & a_{n-1} \\ S\theta_n C\alpha_{n-1} & C\theta_n C\alpha_{n-1} & -S\alpha_{n-1} & -d_n S\alpha_{n-1} \\ S\theta_n S\alpha_{n-1} & C\theta_n S\alpha_{n-1} & C\alpha_{n-1} & d_n C\alpha_{n-1} \\ 0 & 0 & 0 & 1 \end{bmatrix} \quad (1)$$

Where $C\theta_n$ denotes $\cos\theta_n$; $S\theta_n$ denotes $\sin\theta_n$; $S\alpha_{n-1}$ denotes $\sin\alpha_{n-1}$; and $C\alpha_{n-1}$ denotes $\cos\alpha_{n-1}$.

Based on the established coordinate system of the bionic crocodile robot, its D-H parameters were determined as shown in Table Ⅰ:

TABLE I. D-H PARAMETER TABLE OF REAR LIMB

| $i$ | $\theta_i$ /rad | $d_i$ /mm | $a_i$ /mm | $\alpha_i$ /rad |
|---|---|---|---|---|
| 1 | $\theta_1$ | 0 | 70 | 0 |
| 2 | $\theta_2$ | 0 | 0 | $\pi/2$ |
| 3 | $\theta_3$ | 0 | 86 | 0 |
| 4 | $\theta_4$ | 0 | 89 | 0 |

The determined DH parameters are inserted into (1) to obtain the position transformation matrix of each joint of the back leg of the bionic crocodile robot:

$$^0T_1 = \begin{bmatrix} C\theta_1 & -S\theta_1 & 0 & a_1 \\ S\theta_1 & C\theta_1 & 0 & 0 \\ 0 & 0 & 1 & 0 \\ 0 & 0 & 0 & 1 \end{bmatrix} \quad (2)$$

$$^1T_2 = \begin{bmatrix} C\theta_2 & -S\theta_2 & 0 & a_2 \\ 0 & 0 & -1 & 0 \\ S\theta_2 & C\theta_2 & 0 & 0 \\ 0 & 0 & 0 & 1 \end{bmatrix} \quad (3)$$

$$^2T_3 = \begin{bmatrix} C\theta_3 & -S\theta_3 & 0 & a_3 \\ S\theta_3 & C\theta_3 & 0 & 0 \\ 0 & 0 & 1 & 0 \\ 0 & 0 & 0 & 1 \end{bmatrix} \quad (4)$$

$$^3T_4 = \begin{bmatrix} C\theta_4 & -S\theta_4 & 0 & a_4 \\ S\theta_4 & C\theta_4 & 0 & 0 \\ 0 & 0 & 1 & 0 \\ 0 & 0 & 0 & 1 \end{bmatrix} \quad (5)$$

After solving the matrix operation and sorting out, we get (6):

$$^0T_4 = {^0T_1}\,{^1T_2}\,{^2T_3}\,{^3T_4} = \begin{bmatrix} n_x & o_x & a_x & p_x \\ n_y & o_y & a_y & p_y \\ n_z & o_z & a_z & p_z \\ 0 & 0 & 0 & 1 \end{bmatrix} \quad (6)$$

Inserting the parameters from Table I into (6), we obtain::

$$\begin{cases} n_x = C\theta_1 C(\theta_2 + \theta_3 + \theta_4) \\ n_y = S\theta_1 C(\theta_2 + \theta_3 + \theta_4) \\ n_z = S(\theta_2 + \theta_3 + \theta_4) \\ o_x = -C\theta_1 S(\theta_2 + \theta_3 + \theta_4) \\ o_y = -S\theta_1 S(\theta_2 + \theta_3 + \theta_4) \\ o_z = C(\theta_2 + \theta_3 + \theta_4) \\ p_x = a_0 - a_3(C\theta_1 S\theta_2 S\theta_3 - C\theta_1 C\theta_2 C\theta_3) + a_2 C\theta_1 C\theta_2 \\ p_y = S\theta_1(a_3 C(\theta_2 + \theta_3) + a_2 C\theta_2) \\ p_z = a_3 S(\theta_2 + \theta_3) + a_2 S\theta_2 \end{cases} \quad (7)$$

Inverse kinematics involves determining the angles of each joint based on the end-effector's position and orientation. It serves as the foundation for motion planning and foot trajectory control in bio-inspired crocodile robots.

This article employs a geometric analysis method for inverse kinematic analysis. We will position $\theta_1^t$ within the leg schematic of the bio-inspired crocodile robot on the X-Y plane for solving. Fig.7(a) illustrates a schematic diagram of the hind leg mechanism of the bio-inspired crocodile robot in the X-Y plane.

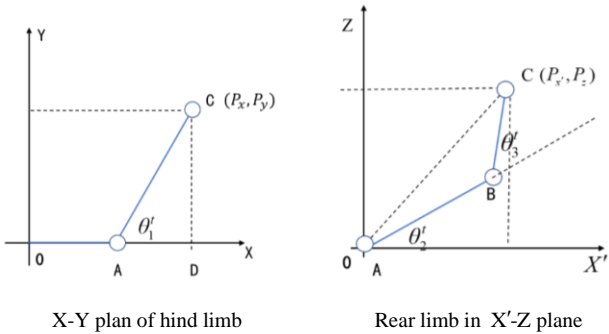

X-Y plan of hind limb       Rear limb in X'-Z plane
Figure.7 A diagram of rear limbs in a plane

Based on the D-H linkage coordinate system established above and the measured data in the three-dimensional diagram of the robot leg, it can be concluded that the leg linkage of the bionic amphibious crocodile robot $OA = a_0^t$. Through the trigonometric function relationship, it can be obtained:

$$\begin{cases} AD = P_x - a_0^t \\ CD = P_y \end{cases} \quad (8)$$

Using formula (8), we can derive the following:

$$\theta_1^t = \arctan \frac{P_y}{P_x - a_0^t} \quad (9)$$

A new coordinate system X'-Z is established on the plane where point ABC is located, and $\theta_2^t$ and $\theta_3^t$ are solved in the schematic diagram of the legs of the bionic crocodile robot on the X'-Z plane. As shown in Fig.7(b), the schematic diagram of the mechanism of the hind legs of the bionic amphibious crocodile robot in the X'-Z plane is shown.

Through the D-H linkage coordinate system established above and the data measured by the three-dimensional diagram of the robot leg, it can be obtained that the leg linkage $AB = a_2^t$, the leg linkage $BC = a_3^t$, and the foot coordinate $(\sqrt{(P_x - a_0^t)^2 + P_y^2}, P_z)$, where O point is the coordinate origin. Where $\theta_3^t$ is an outer Angle of triangle ABC, it can be obtained through the trigonometric function relationship:

$$\begin{cases} AB = a_2^t \\ BC = a_3^t \\ AC = \sqrt{(P_x - a_0^t)^2 + P_y^2 + P_z^2} \\ \angle ABC = \arccos\left[\dfrac{AB^2 + BC^2 - AC^2}{2AB \times AC}\right] \end{cases} \quad (10)$$

According to the formula (10), it can be obtained:

$$\theta_3^t = 180° - \angle ABC = 180° - \arccos\left[\frac{a_2^{t2} + a_3^{t2} - ((P_x - a_0^t)^2 + P_y^2 + P_z^2)}{2a_2^t \times a_3^t}\right] \quad (11)$$

$$\begin{cases} \angle CAZ = \arctan\left[\dfrac{\sqrt{(P_x - a_0^t)^2 + P_y^2}}{P_z}\right] \\ \angle CAB = \arccos\left[\dfrac{a_2^{t2} - a_3^{t2} + ((P_x - a_0^t)^2 + P_y^2 + P_z^2)}{2a_2^t \times \sqrt{(P_x - a_0^t)^2 + P_y^2 + P_z^2}}\right] \\ \theta_2^t = 90° - \angle CAZ - \angle CAB \end{cases} \quad (12)$$

$$\theta_2^t = 90° - \arctan\left[\frac{\sqrt{(P_x - a_0^t)^2 + P_y^2}}{P_z}\right] - \arccos\left[\frac{a_2^{t2} - a_3^{t2} + ((P_x - a_0^t)^2 + P_y^2 + P_z^2)}{2a_2^t \times \sqrt{(P_x - a_0^t)^2 + P_y^2 + P_z^2}}\right] \quad (13)$$

The inverse kinematics solution of the legs of the bionic crocodile robot is as follows:

$$\begin{cases} \theta_1^t = \arctan\dfrac{P_y}{P_x - a_0^t} \\ \theta_2^t = 90° - \arctan\left[\dfrac{\sqrt{(P_x - a_0^t)^2 + P_y^2}}{P_z}\right] - \arccos\left[\dfrac{a_2'^2 - a_3'^2 + ((P_x - a_0^t)^2 + P_y^2 + P_z^2)}{2a_2' \times \sqrt{(P_x - a_0^t)^2 + P_y^2 + P_z^2}}\right] \\ \theta_3^t = 180° - \angle ABC = 180° - \arccos\left[\dfrac{a_2'^2 + a_3'^2 - ((P_x - a_0^t)^2 + P_y^2 + P_z^2)}{2a_2' \times a_3'}\right] \\ \theta_4^t = \theta_4^t \end{cases}$$

(14)

## B. Kinematic model of spine

The D-H linkage coordinate system of the spine of the bionic crocodile robot is shown in Fig.8.

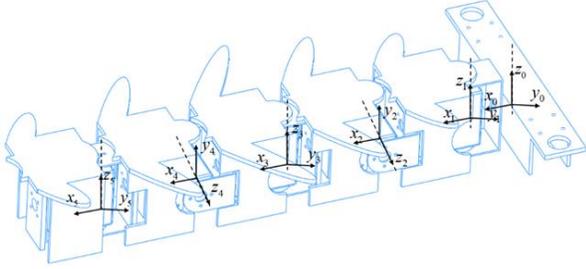

Figure.8 The D-H coordinate system of the torso

Based on the established D-H coordinate system of the torso, the relevant D-H parameters were determined as shown in Table Ⅱ:

TABLE II. D-H PARAMETERS OF THE TORSO

| $i$ | $\theta_i$ /rad | $d_i$ /mm | $a_i$ /mm | $\alpha_i$ /rad |
|---|---|---|---|---|
| 1 | $\theta_1$ | 0 | 50.5 | 0 |
| 2 | $\theta_2$ | 0 | 60.5 | $-\pi/2$ |
| 3 | $\theta_3$ | 0 | 60.5 | $\pi/2$ |
| 4 | $\theta_4$ | 0 | 60.5 | $-\pi/2$ |
| 5 | $\theta_5$ | 0 | 60.5 | $\pi/2$ |

After solving the matrix operations and organizing them, we obtain the following:

$$\begin{cases} P_x = 60.5(C\theta_1 + C\theta_1 C\theta_2 - S\theta_1 S\theta_3 - C\theta_4 S\theta_1 S\theta_3 - C\theta_1 S\theta_2 S\theta_4 + C\theta_1 C\theta_2 C\theta_3) + 0.5C\theta_1 C\theta_2 C\theta_3 + 50.5 \\ P_y = 60.5(S\theta_1 + S\theta_1 C\theta_2 + C\theta_1 S\theta_3 + C\theta_1 C\theta_4 S\theta_3 - S\theta_1 S\theta_2 S\theta_4 + C\theta_2 C\theta_3 S\theta_1) + 0.5C\theta_2 C\theta_3 S\theta_1 \\ P_z = 60.5(-S\theta_2 - S\theta_2 C\theta_3 - C\theta_2 S\theta_4 - C\theta_3 C\theta_4 S\theta_2) \end{cases}$$

(15)

## C. Analysis of flexible tail

The tail section of the bionic crocodile robot designed in this paper is shown in Fig.9, which mainly contains three parts: drive, line drive flexible and underdrive section.

As shown in Fig.9, the XOY coordinate system is set at the center of the joint plate between the drive part and the wire-driven flexible part. The flexible wire part consists of six joints of equal length, driven by a servomotor through a pair of inextensible wire cords of length L. When the servo motor rotates, one cord is elongated, and the other is shortened, prompting the overall bending of the tail.

The angle of curvature $\theta$ of the flexible part of the wire driving is related to the length variation $X$ of the wire rope, and the length variation $X$ of the wire rope is related to the angle of rotation $\phi$ of the servo motor.

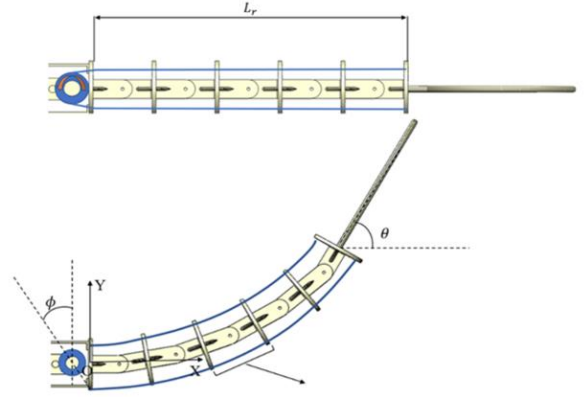

Figure.9 Design of the flexible tail of the bionic crocodile robot

The angle of rotation of each joint is the same, and the change in length of the two strings can be obtained as follows:

$$\begin{cases} \Delta h_D = -\left[d\sin\left(\dfrac{\theta}{2N}\right) + 2h\sin^2\left(\dfrac{\theta}{4N}\right)\right] \\ \Delta h_C = d\sin\left(\dfrac{\theta}{2N}\right) - 2h\sin^2\left(\dfrac{\theta}{4N}\right) \end{cases}$$

(16)

Where $N$ is the number of joints, $\Delta h_D$ is the length of the shortened cord, and $\Delta h_C$ is the length of the extended cord. To simplify the calculation, the quadratic term in (16) can be ignored because the angle of rotation $\dfrac{\theta}{N}$ of each joint is very small.

$$\begin{cases} \Delta h_D = -d\sin\left(\dfrac{\theta}{2N}\right) \\ \Delta h_C = d\sin\left(\dfrac{\theta}{2N}\right) \end{cases}$$

(17)

The relationship between the angle of rotation of the servomotor and the length of the string extended or shortened is shown as follows, where $r$ is the radius of the steering wheel of the drive motor, $\phi$ is the rotation Angle of the steering wheel, $N$ is the number of joints of the flexible tail, and $\theta$ is the final deviation Angle of the flexible tail of bionic crocodile robot.

$$\phi = \dfrac{180}{\pi r} Nd \sin\left(\dfrac{\theta}{2N}\right)$$

(18)

The position of each joint in the XOY coordinate system can be expressed as:

$$\begin{cases} x_i = \begin{cases} \dfrac{H+h}{2}; i=1 \\ \sum_{j=1}^{i-1}(H+h)\cos(\dfrac{i\theta}{N}); i \geq 2 \end{cases} \\ y_i = \begin{cases} 0; i=1 \\ \sum_{j=1}^{i-1}(H+h)\cos(\dfrac{i\theta}{N}); i \geq 2 \end{cases} \end{cases}$$
(19)

where $(x_i, y_i)$ denotes the position of the $i$ th joint in the coordinate system XOY and $N$ is the number of active joints in the tail of the line drive.

*D. Motion analysis of coordinated planning for multi parts*

Crocodiles exhibit an extraordinary degree of flexibility and adaptability in their natural environment, demonstrating remarkable freedom of movement. This exceptional mobility is achieved through the coordinated cooperation of their limbs and other body parts, including the spine and tail. This underscores the critical role of synergy among these body components. Therefore, gaining a deeper understanding of the coordination mechanisms in crocodile locomotion is highly relevant for enhancing the motion control of bio-inspired crocodile robots.

By cultivating self-awareness in the limbs, spine, and tail, and by closely observing the relative positions of these body parts during various movements such as crawling and jumping, we can achieve precise control over each aspect of the bio-inspired crocodile. Simultaneously, we can effectively coordinate the control of interactions among these components.

We have established three feedback rules by utilizing self-perception feedback to achieve closed-loop control for individual components. These rules are as follows:

(1) Limb-to-limb perception feedback.
(2) Spine-to-spine perception feedback.
(3) Tail-to-tail perception feedback.

Through these three rules, we can achieve stable closed-loop control of the limbs, spine, and tail of the bio-inspired crocodile robot. While the controllers for the limbs, spine, and tail operate independently, they are not isolated; they share certain variables (such as limb phase, spine bending angles and directions, tail bending angles and directions) to ensure overall temporal and spatial coordination.

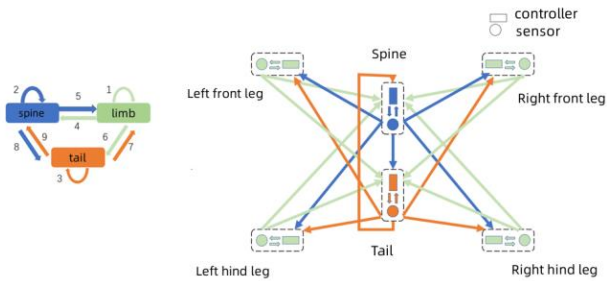

Figure.10 Topological model

Fig.10 presents a topological model established based on the biomechanical analysis of crocodile movements. In the model, we have incorporated foot-end pressure sensors, as well as self-torque and position feedback sensors for the servo motors. We employ PID controllers for closed-loop control of the joint angles in the bio-inspired crocodile. The robot's leg section comprises one yaw joint servo motor, two pitch joint servo motors, and a joint angle controller. The spine section includes three yaw joint servo motors, two pitch joint servo motors, and a joint angle controller. The tail joint section consists of one pitch joint servo motor, a linear actuator for tail movement, and a joint angle controller. The joint angle controller is responsible for outputting the angle $\theta_i$ for each driver, where $\theta_i$ represents the target angle for each joint. In the target angle function, $i$ represents the driver motor for each joint.

The video of the crocodile's movement was decomposed into continuous 3D snapshots using ScreenToGif software. The relationship between the angle of the spine, the angle of the tail and the phase of the crocodile's limbs was explored to link the various parts of the crocodile's body. The definitions of the angles of the spine and tail are shown in Fig.12.

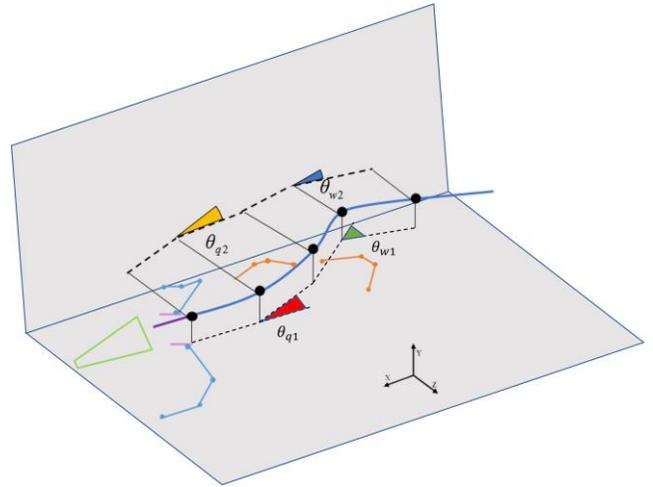

Figure.11 Definition of angle and direction of crocodile spine and tail

The black dots in Fig.11 indicate the points used to calculate the angles, and the corresponding line segments are indicated by black dashed lines. The angles are shown as red and green triangular areas (for pitch angles in the x-z plane) and yellow and blue triangular areas (for yaw angles in the x-y plane). Where $\theta_{q1}$ indicates the pitch angle of the spine of the bionic crocodile robot. The angle is defined as negative when the spine is bent to the left side of the crocodile body and positive when the spine is bent to the right side of the robot body. $\theta_{w1}$ is defined as the pitch angle of the tail of the robot. The angle is defined as negative when the tail is bent to the left side of the robot body. $\theta_{q2}$ is defined as the yaw angle of the spine of the robot. The angle is defined as negative when the spine is bent toward the top of the robot's body. $\theta_{w2}$ is defined as the yaw angle of the tail of the robot. The angle is defined as negative when the tail is bent to the top of the robot body and positive when the tail is bent to the bottom of the robot body.

For the example of the crocodilian crawl movement, the corresponding relationship between the angle of the spine joint, the angle of the tail joint, and the gaits in two cycles was extracted and obtained by the above procedure, as shown in Fig.12.Among them, the black area in the limb phase represents the landing phase, and the black area in the plots of the tail and spine represents the positive angle, LQ, RQ, LH and RH respectively refer to the left front leg, right front leg, left hind leg and right hind leg..

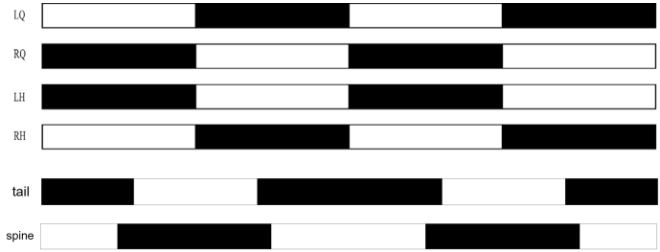

Figure.12 The correspondence between the angle of the spine, the angle of the tail and the phase of the gait.

IV. RESULTS

*A. Motion stability*

Two groups of bionic crocodile robots have been simulated for testing whether the coordinated movement of all parts of the bionic crocodile robot can enhance its stability of motion.

Under the condition of unique variables, a group of bionic crocodile robots were set to participate in the movement of their limbs, tail and spine when crawling, that is, the tail dragged the ground and the tail and spine swung in the left and right direction, while the other group of robots only participated in the movement of their limbs when crawling, the tail lifted off the ground, and the spine and tail were fixed. The changes in the center of gravity of the two groups of bionic crocodile robots were recorded respectively.

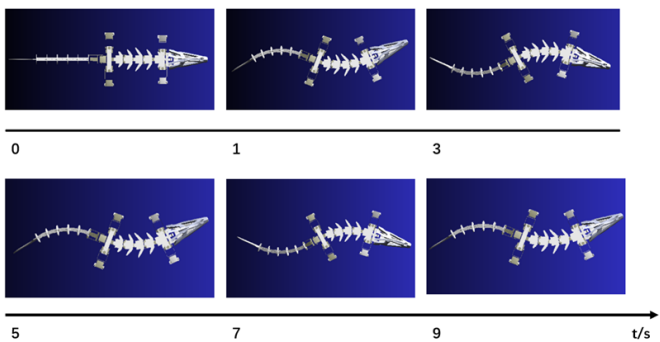

Figure.13 Movement gait of the bionic crocodile robot during crawling

The simulation experiment of the bionic crocodile robot is shown in Fig.13. The experiment contains two complete crawling cycles, where 1-5 seconds is the first crawling cycle and 5-9 seconds is the second crawling cycle.

The state of the bionic crocodile robot during the crawling period was observed at three selected time points within a complete crawling cycle, as shown in Fig.14, which are the screenshots of the simulation experiments at 1.2s, 1.5s and 1.8s, respectively.

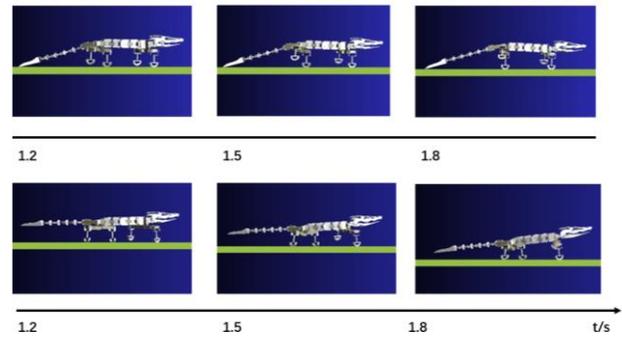

Figure.14 Comparison of the stability of a bionic crocodile robot during crawling

The upper part of Fig.14 shows the motion state of the bionic crocodile robot with the participation of the tail, body and limbs, and it crawls forward with a diagonal gait [23][24]. During the motion of the robot, the spine swings from side to side in a sinusoidal pattern with the limbs, and the tail stays on the ground and swings from side to side to provide support for the body and improve the stability of the body. The bottom of Fig.14 shows the crawling state when the tail and spine are not involved in the motion. When the tail of the robot the robot's tail is not involved in the movement and it moves in a diagonal gait, the two legs on the ground take on the role of supporting and driving the body forward at the same time. The center of gravity should always be on the line of the two supporting points in order to achieve stability. The simulation experiment shows that at 1.2 seconds, the bionic crocodile robot tilts its body because the center of gravity is not on the line of the support points, and the robot shifts toward the center of gravity.

Fig. 15 shows the change in the height of the center of gravity during the movement of the two groups of bionic crocodile robots, where the orange and blue represent the change in the height of the center of gravity with and without the tail involved in the movement, respectively. From the figure, we see that the change in height of the center of gravity of the bionic crocodile robot varies in a certain way in both simulation experiments, but the variation amplitude of the set with the tails involved in the movement is significantly smaller than that of the set without the tails involved in the movement, showing that the tails involved in the movement of the bionic crocodile robot are more stable.

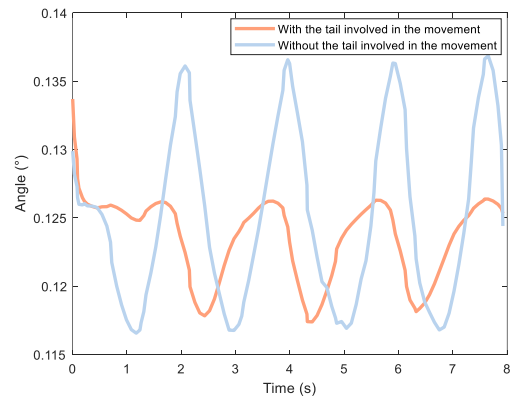

Figure.15 Comparison of the fluctuation of the center of gravity between the tail participating in movement and the tail not participating in movement

### B. The effect of tail and spine swing on movement

In order to study the influence of the tail and spine of the bionic crocodile robot on the motion, the corresponding land motion simulation experiment and the underwater motion simulation experiment were set up respectively.

In order to study the function of the tail and spine of the bionic crocodile robot when moving on land, a comparative simulation experiment was set up. The displacement of the two groups of robots moving on land is shown in Figure 16.

Figure 16 depicts the forward displacement of the bionic crocodile robot. The orange and blue curves represent the displacement with or without the spine and tail involved in the movement, respectively. The results indicate that within the same simulation time, the robot's forward displacement is 0.16 meters when the spine and tail are not engaged in the motion. However, when the spine, tail, and limbs are involved in the motion simultaneously, the robot achieves a displacement of 0.73 meters in the forward direction. This comparison demonstrates that when the spine and tail are engaged in the movement, the robot attains a higher velocity, covering more distance in the same time.

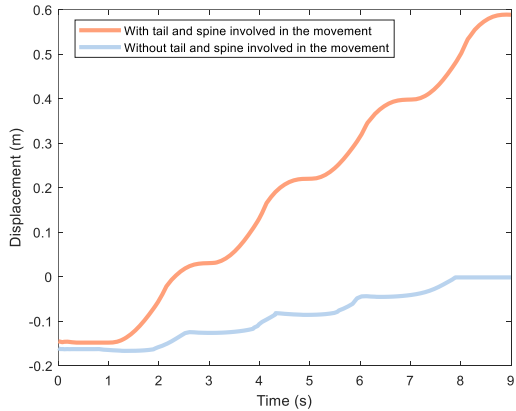

Figure.16 Displacement with and without spine and tail involved in movement

Upon observing the swimming behavior of real crocodiles in the water, it becomes evident that the primary forward thrust generated during crocodile swimming is achieved through the coordinated movement of the crocodile's body and the powerful action of its tail. To verify the role of the crocodile's spine and tail during aquatic locomotion, FLUENT software was employed to analyze the swimming dynamics of the bionic crocodile.

The swimming pattern of crocodiles in water closely resembles that of fish. Consequently, the body and tail movements of crocodiles are described in this paper as fish-like body waves. The fish body wave curve is represented as a combination of the amplitude envelope of the fish body and a sine curve. The mathematical function for the fish body wave is as follows:

$$y_{body}(x,t) = (c_1 x + c_2 x^2)\sin(kx + \omega t) \quad (20)$$

By measuring the center line parameters of the bionic crocodile robot, the center line trajectory is drawn and modeled into the form of fish body wave function. The parameters of the fish body wave curve are obtained as $c_1 = 0.027$, $c_2 = 0.30$ and $k = 0.023$.

As crocodiles predominantly employ trunk and tail movements for propulsion during swimming, this paper simplifies the bionic crocodile model by omitting the limbs and focuses on a two-dimensional body for simulation analysis.

In Figure 17, the pressure distribution is visualized during the swimming motion of the bionic crocodile robot. Figure 17(a) specifically displays the pressure distribution at T/4. It is apparent that the pressure on the bending and convex side of the flexible tail is greater than that on the concave side. This pressure differential between the two sides of the flexible tail results in a diagonal forward force, propelling the robot in a diagonal direction. As the robot's body and tail continue their movement towards the head-to-tail axis, the pressure difference gradually decreases, as demonstrated in Figure 17(c). When the bionic crocodile's torso and tail reach 3T/4, the pressure difference between the two sides of the flexible tail generates a symmetrical diagonal forward thrust, producing an "S"-shaped swimming trajectory, closely resembling the natural swimming motion of real crocodiles.

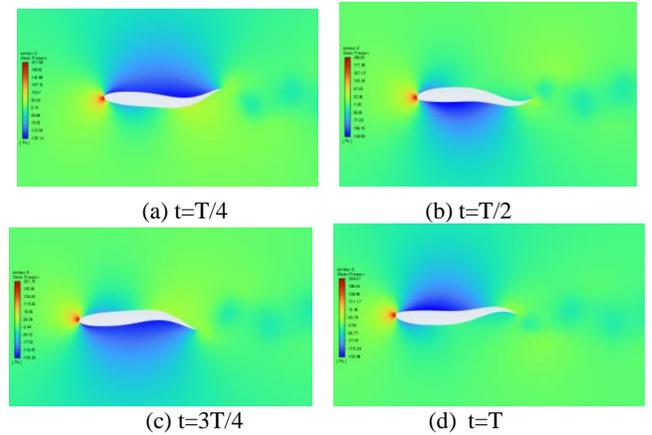

(a) t=T/4    (b) t=T/2

(c) t=3T/4    (d) t=T

Figure.17 Pressure nephogram of bionic crocodile robot in one swimming cycle

Two sets of experiments show that when the bionic crocodile robot moves on land, the swing of the tail and spine can improve the motion displacement; When moving through the water, the swing of the tail and spine provides the main thrust.

### C. Motion fault-tolerance

In order to study the fault tolerance of the bionic crocodile robot in the coordinated movement of multiple parts, we have assumed a paralysis of the leg in the experiment and compared its fault tolerance rate. Joint damage indicates that the actuator of the joint is unable to receive and execute the motion commands. Here, four sets of comparative tests are set up to study the fault tolerance of the bionic crocodile robot when the front leg rotational joint, the rear leg rotational joint, the front leg pitching joint, and the rear leg pitching joint are paralyzed, respectively.

Fig.18 shows the displacement with and without the involvement of the spine in the case of a damaged rotating joint of the front leg. As can be seen from the figure, compared to the robot with intact joints and under coordinated control of multiple parts, the robot with damaged rotating joints of front leg and without trunk cooperative motion has an 80% reduction in locomotion, and the robot with damaged front leg rotating joints but with trunk cooperative motion has only a 22% reduction in locomotion.

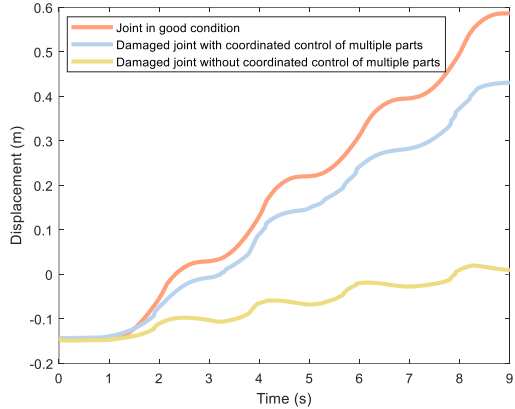

Figure.18 Displacement of different bionic crocodile robots in the absence of rotating joints of the front leg

Fig.19 shows the displacement with and without spine involvement motion in the case of a damaged hind leg rotational joint. As can be seen from the figure, compared to the robot with intact joints and under coordinated control of multiple parts, the robot with a damaged hind leg rotating joint and without trunk cooperative motion has a 75% decrease in locomotion, while the robot with a damaged hind leg rotating joint but with trunk cooperative motion has only a 17% decrease in locomotion.

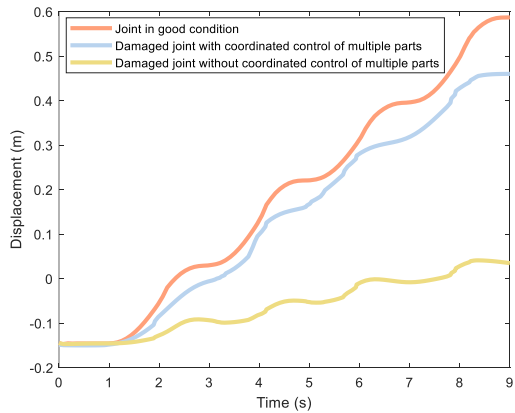

Figure.19 Displacement of different bionic crocodile robots in the absence of rotating joints of the hind leg

Fig.20 shows the displacement with and without the involvement of the spine in the case of damage to the pitching joint of the front leg. From the figure, it can be seen that the damage to the pitching joint of the front leg caused the bionic crocodile robot to lose the ability of forward motion.

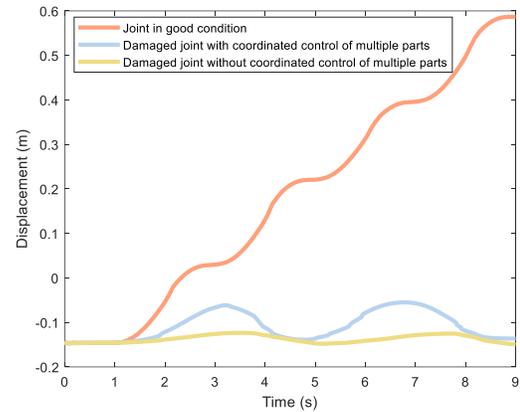

Figure.20 Displacement of different bionic crocodile robots in the absence of pitching joint of front leg

Fig.21 shows the displacement with and without the spine involved in the motion in the case of a damaged pitching joint of the hind leg. As can be seen from the figure, when the hind legs of the robot are missing, the bionic crocodile robot coordinated by multiple parts raises the center of gravity of the robot through the support function of the tail, right front leg and left back leg, so that the hind legs can be lifted off the ground and swing. Moreover, the ground where the robot moves is relatively flat and does not require excessive step height, so the absence of pitching joint of the hind legs can be made up through the coordinated control of multiple parts. The robot with a damaged pitching joint of hind legs and without cooperating motion of trunk had a 55% reduction in locomotion velocity due to the lack of torso coordination. In contrast, the robot with a damaged pitching joint of hind legs and with cooperating motion of trunk showed almost no reduction in locomotion compared to the robot with an intact joint.

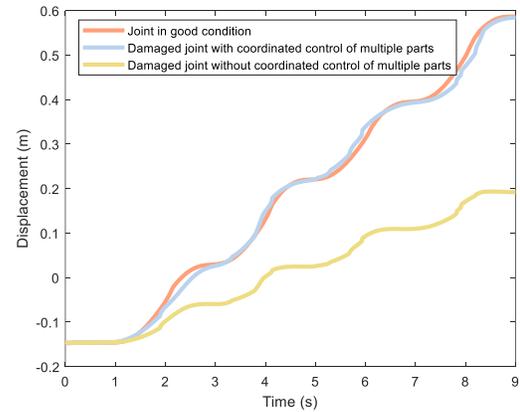

Figure.21 Displacement of different bionic crocodile robots in the absence of pitching joint of hind leg

*D. Motion adaptability*

As vertebrate reptiles, crocodiles have a low center of gravity when crawling on land, and they have shorter legs and lower reachable heights. However, the crocodile is more adaptable to the external environment, and it can use its strong legs and powerful tail to achieve a semi-standing posture, which increases its reachable height and facilitates its predation of prey at high places. To ensure that the robot studied in this paper is highly biomimetic, the spine angle, tail angle and foot phase are extracted from a real crocodile standing bipedally, and the relationship between them is established.

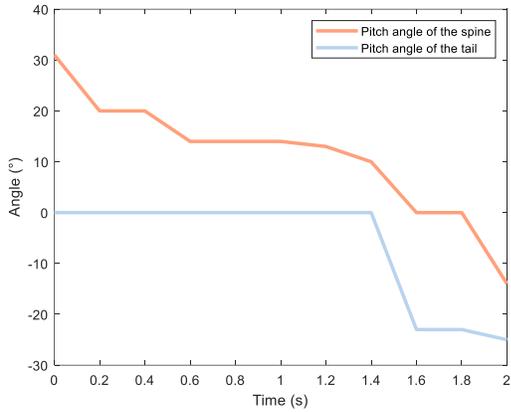

Figure.22 Changes in the angle of the spine and tail of the crocodile standing on two legs

Fig.23 shows the side view of the bionic crocodile robot in the crawling state and the bipedal standing state. From the figure, it can be seen that the reachable height of the bionic crocodile robot depends on the length of its legs when crawling on land. When the bionic crocodile robot stands through the biped and tail with the cooperation of the spine, tail and hind limbs, the reachable height is determined by the length of the spine, the length of the head and the length of the hind legs together.

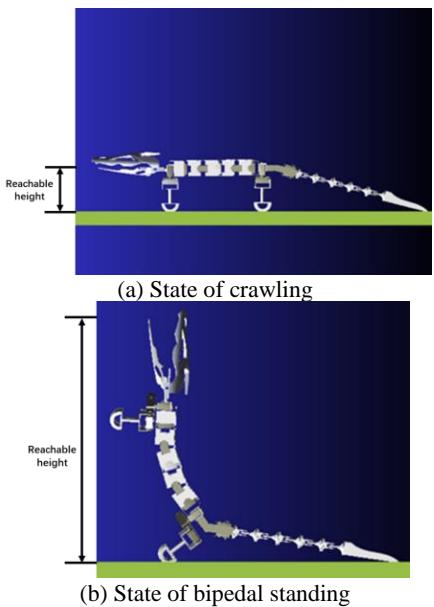

(a) State of crawling

(b) State of bipedal standing

Figure.23 Bionic crocodile robot stands bipedally

Fig.24 shows the change in the height that the bionic crocodile robot can reach during the process from the quadruped to the biped standing condition. As shown in the figure, the bionic crocodile robot can reach a height of 0.16m in the crawling state and 0.68m in the bipedal standing state.

The reachable height in the bipedal standing state is 4.25 times higher than that in the crawling state.

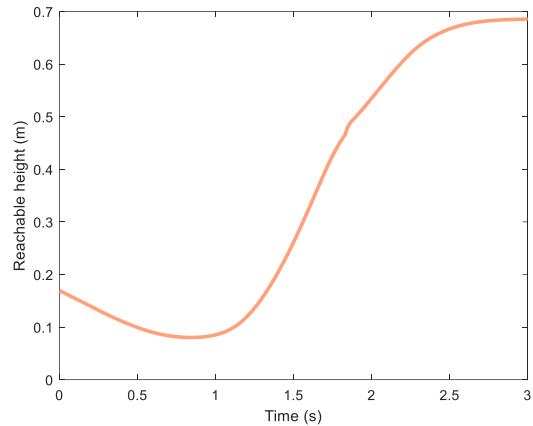

Figure.24 Variation of reachable height for bipedal standing of a bionic crocodile robot

The experiments show that the reachable height of the bionic crocodile robot is limited due to the constraint of leg length when crawling on land. The coordination between the tail, spine and hind limbs can improve the reachable height of the crocodile, thus the working space of the crocodile robot can be greatly improved and the adaptability to the environment can be enhanced.

V. CONCLUSION

In this paper, we focus on the bionic crocodile robot, using the crocodile as our biological model. Firstly, we design the structure of the bionic crocodile robot based on the morphological features and motion characteristics of real crocodiles. We employ the D-H method to establish kinematic models for the robot's legs and spine. Subsequently, we conduct experimental analyses to evaluate the motion stability of the bionic crocodile robot under coordinated control of its spine, tail, and limbs, assessing its robustness in the face of potential damage and its ability to adapt to varying environments.

The results of our simulation experiments indicate that involving the spine and tail in conjunction with the limbs enhances the stability of the bionic crocodile robot during movement. This combination results in increased displacement, heightened fault tolerance when individual joints are compromised, and improved adaptability to different environmental conditions. Consequently, we can deduce that limb coordination plays a pivotal role in animal locomotion.

Based on our theoretical analysis and the outcomes of the simulation experiments, we have verified that the bionic crocodile robot, as designed in this study, can attain a larger range of motion and fulfill a broader spectrum of functions when its components work together in a coordinated manner.

REFERENCES

[1] Akira F, Megu G, Yoichi M. Comparative anatomy of quadruped robots and animals: a review[J]. Advanced Robotics, 2022,36(13).
[2] A Quadruped Robot Exhibiting Spontaneous Gait Transitions from Walking to Trotting to Galloping.

[3] Hildebrand M 1959 Motions of the running cheetah and horse J. Mammal. 40 481–95.

[4] JagnandanK and Timothy E H 2017 Lateral movements of a massive tail influence gecko locomotion: an integrative study comparing tail restriction and autotomy Sci. Rep. 7.1 10865.

[5] Jia X, Frenger M, Chen Z, et al. An alligator inspired modular robot[J]. Proceedings - IEEE International Conference on Robotics and Automation, 2015,2015:1949-1954.

[6] Jia X, Chen Z, Petrosino J M, et al. Biological undulation inspired swimming robot[C]//2017 IEEE International Conference on Robotics and Automation (ICRA). IEEE, 2017: 4795-4800.

[7] Karwa K G, Mondal S, Kumar A, et al. An open source low-cost alligator-inspired robotic research platform[C]//2016 Sixth International Symposium on Embedded Computing and System Design (ISED). IEEE, 2016: 234-238.

[8] Agrawal K, Jain K, Gupta D, et al. Bayesian Optimization Based Terrestrial Gait Tuning for a 12-DOF Alligator-Inspired Robot With Active Body Undulation[C]//International Design Engineering Technical Conferences and Computers and Information in Engineering Conference. American Society of Mechanical Engineers, 2018, 51807: V05AT07A076.

[9] Godiyal R, Zodage T, Rane T. 2DxoPod-A Modular Robot for Mimicking Locomotion in Vertebrates[J]. Journal of Intelligent & Robotic Systems, 2021, 101(1): 1-16.

[10] Brochu C A. Phylogenetic Approaches Toward Crocodylian History[J]. Annual Review of Earth and Planetary Sciences, 2003,31(1):357-397.

[11] Brochu C A. Phylogenetics, taxonomy, and historical biogeography of Alligatoroidea. In Cranial morphology of Alligator mississippiensis and phylogeny of Alligatoroidea. Edited by T. Rowe, C.A. Brochu, and K. Koshi. Society of Vertebrate Paleontology, Memoir 6[J]. Journal of Vertebrate Paleontology, 1999,19:9-100.

[12] Straub, Carol S. "Crocodiles; Their Natural History, Folklore and Conservation." (1973): 404- 405.

[13] Harshman J, Huddleston C J, Bollback J P, et al. True and false gharials: a nuclear gene phylogeny of crocodylia[J]. Systematic Biology, 2003,52(3):386-402.

[14] Webb G J, Manolis S C, Brien M L. Saltwater crocodile Crocodylus porosus[J]. Crocodiles— Status Survey and Conservation Action Plan. Third Edition. Darwin: Crocodile Specialist Group, 2010:99-113.

[15] 丁由中, 王小明, 何利军, 等. 野生扬子鳄种群及栖息地现状研究[J]. 生物多样性, 2001,9(2):102.

[16] Grigg G, Gans C. Morphology and physiology of the Crocodylia[J]. Fauna of Australia, 1993, 2: 326-336.

[17] Fish F E. Kinematics of undulatory swimming in the American alligator[J]. Copeia, 1984: 839- 843.

[18] Wiseman A L A, Bishop P J, Demuth O E, et al. Musculoskeletal modelling of the Nile crocodile (Crocodylus niloticus) hindlimb: effects of limb posture on leverage during terrestrial locomotion[J]. Journal of anatomy, 2021, 239(2): 424-444.

[19] Hutton J M. Morphometrics and field estimation of the size of the Nile crocodile[J]. African Journal of Ecology, 1987,25(4):225-230.

[20] Willey J S, Biknevicius A R, Reilly S M, et al. The tale of the tail: limb function and locomotor mechanics in Alligator mississippiensis[J]. Journal of experimental biology, 2004, 207(3): 553- 563.

[21] Veeger H E J, van der Helm F C T. Shoulder function: The perfect compromise between mobility and stability[J]. Journal of Biomechanics, 2007,40(10):2119-2129.

[22] wa T. Foundations of robotics: analysis and control[M]. MIT press, 1990.

[23] Milton H. The Adaptive Significance of Tetrapod Gait Selection[J]. American Zoologist, 1980,20(1).

[24] Wassersug R J. Tetrapod Gait Patterns[J]. Science, 1975,188(4195)